\documentclass[10pt,journal]{IEEEtran}

\usepackage{float}
\usepackage{url}

\ifCLASSINFOpdf
  \usepackage[pdftex]{graphicx}
  \graphicspath{{./figures/}}
  \DeclareGraphicsExtensions{.pdf,.jpeg,.png}
\else
  \usepackage[dvips]{graphicx}
  \graphicspath{{./figures/}}
  \DeclareGraphicsExtensions{.eps}
\fi

\usepackage{amsmath}
\usepackage{amsfonts}
\usepackage{siunitx}
\newrobustcmd{\B}{\fontseries{b}\selectfont}
\usepackage{multirow}
\usepackage{pifont}
\usepackage{tabularx}
\usepackage{subcaption}

\usepackage{verbatim}
\usepackage[numbers]{natbib}
\begin{document}

\title{Open-Set: ID Card Presentation Attack Detection using Neural Transfer Style}

\author{Reuben~Markham, 
         Juan~M.~Espín, 
         Mario~Nieto-Hidalgo,
         Juan~E.~Tapia,~\IEEEmembership{Member,~IEEE}
 \thanks{Reuben Markham, Instituto Tecnológico de Informática (ITI), Universitat Politècnica de València, 46022 Valencia, Spain, e-mail: (rmarkham@iti.es).}
 \thanks{Juan Manuel Espín, Facephi Biometria SA, 03003 Alicante, Spain, e-mail: (jmespin@facephi.com).}
 \thanks{Mario Nieto-Hidalgo, Facephi Biometria SA, 03003 Alicante, Spain, e-mail: (marionieto@facephi.com).}
 \thanks{Corresponding author: Juan Tapia, da/sec-Biometrics and Internet Security Research Group,
 Hochschule Darmstadt, Germany, e-mail: (juan.tapia-farias@h-da.de)}
 \\{\textbf{** This publication is a pre-print version **}}
}




\maketitle


\begin{abstract}
The accurate detection of ID card Presentation Attacks (PA) is becoming increasingly important due to the rising number of online/remote services that require the presentation of digital photographs of ID cards for digital onboarding or authentication. Furthermore, cybercriminals are continuously searching for innovative ways to fool authentication systems to gain unauthorized access to these services. Although advances in neural network design and training have pushed image classification to the state of the art, one of the main challenges faced by the development of fraud detection systems is the curation of representative datasets for training and evaluation.
The handcrafted creation of representative presentation attack samples often requires expertise and is very time-consuming, thus an automatic process of obtaining high-quality data is highly desirable. This work explores ID card Presentation Attack Instruments (PAI) in order to improve the generation of samples with four Generative Adversarial Networks (GANs) based image translation models and analyses the effectiveness of the generated data for training fraud detection systems. Using open-source data, we show that synthetic attack presentations are an adequate complement for additional real attack presentations, where we obtain an EER  performance increase of 0.63 \% points for print attacks and a loss of 0.29 \% for screen capture attacks.
\end{abstract}

\section{Introduction}

\IEEEPARstart{I}{n} recent years, a growing trend to digitalise processes that traditionally required physical attendance and the presentation of official ID documents has been observed. This tendency has been driven mostly by technological advances, novel legal regulations, and also pandemics, where long-term confinements force to look for alternatives to traditional ways of accessing certain services to remote services. Examples of affected processes are opening accounts in financial institutions, logistics, transport, retail or asset investment platforms, taking out insurance, and purchasing real estate \cite{gonzalez2021hybrid}. 

The proliferation of remote services that require the identification of natural persons through biometrics and ID documents has motivated the continued search for weaknesses in the said process by the attacker in order to access the services without being identified. A common strategy is the presentation of documents that have been digitally manipulated and then printed on glossy or bond papers and represented on a smartphone or tablet screen, also known as spoofs or Presentation Attacks (PA).

The increasing sophistication and effectiveness of the methodologies with which attackers create convincing fake documents highlights the need to develop increasingly effective ID document Presentation Attack Detection (PAD) systems. These systems have, as an essential component, an image classifier to distinguish between bona fide documents and PA. The current trend for creating image classifiers is to use neural network architectures, specifically Convolutional Neural Networks (CNN) \cite{lecun1989backpropagation} or Vision-Transformers (VT) \cite{vaswani2017attention}, and to train them to minimize classification errors.

Moreover, it is well known that the stable training of modern neural networks requires a large set of diverse data to reach generalization capabilities. In the context of ID documents, data acquisition is a significant challenge because the data is subject to privacy concerns and legal regulations such as the GDPR\footnote{\url{https://gdpr-info.eu/}}, which requires the consent of the subject for the processing and use of their data. Furthermore, the obtainment of PA would involve the laborious process of printing and cutting out of documents or preparing screen presentations with different display monitors \cite{Tapia-PS}. The difficulties associated with the procurement of data have limited the quality and quantity of public research associated with the development of PAD models since the studies must often rely on in-house or private datasets, which makes it impossible to replicate the reported results (See Table \ref{tab:dataset_characteristics_studies}). To alleviate these deficiencies and promote innovation, public datasets of synthetic documents have appeared in recent years, notably the Mobile Identity Document Video (MIDV) datasets \cite{midv500,midv2019,midv2020}, the Document Liveness Challenge (DLC-2021) dataset \cite{dlc2021}, and the Synthetic Chilean ID Card dataset \cite{benalcazar2023synthetic}.

This work leverages open-source datasets of video clips containing presentations of ID documents of fake subjects with the aim of ascertaining whether augmenting the training set with synthetic presentation attack samples instead of bona fide samples yields comparable results in terms of PAD predictive performance. To that end, two tasks are specified: firstly, the ``print'' task, where the model must distinguish between bona fide and print attack species, and lastly, the ``screen'' task, where bona fide and screen presentations are discriminated by the system. The datasets constructed for both tasks comprise of preprocessed frames of the original clips, where each image is a presentation of a full, aligned document with background information removed. Thus, the three species considered in this work for classification are bona fide, print and screen:
\begin{itemize}
    \item \textit{Bona fide}: Video clips of ID cards containing synthetic data were captured in a variety of situations with smartphone cameras.  
    \item \textit{Print}: Digital templates of ID cards were printed on normal paper and cut out. Then, smartphones were used to capture short clips of the printed cards in different situations.
    \item \textit{Screen}: Templates of ID cards were shown on computer and tablet screens, after which a smartphone was used to capture clips of the depicted images. 
\end{itemize}
Supervised and unsupervised image-to-image translation models based on Generative Adversarial Networks (GANs) are explored to increase the number of presentation attack samples in the training dataset. This work is heavily inspired by a recent study \cite{benalcazar2023synthetic} that employs GANs and texture transfer-based algorithms to generate bona fide and presentation attack samples. In the same vein as the aforementioned work, the usefulness of the generated images is assessed by training several MobileNetV2 \cite{sandler2018mobilenetv2} networks for each task. This way, the predictive performance using training sets comprised of synthetic and real samples can be compared with that of systems obtained by training with only real data.

In summary, the main contributions of this paper are:

\begin{itemize}
\item This work proposes a comprehensive analysis of the State-Of-The-Art related to PAD on ID cards and open-access databases. 
\item The GAN-based methods are explored in order to generate synthetic images to simulate and replicate the print and screen PA. This reduces the time to produce handcrafted attacks.
\item Supervised and unsupervised presentation attack generation methods based on GANs are developed from supervised and unsupervised data for generating PA of full ID cards that retain the content of the original bona fide images.
\item The system is trained using only open-access databases instead of private images used in the SOTA (Not available). Then, we show the improvements, limitations and tangible results we can reach with the open-access databases.
\item This work shows and highlights the competitive results of developing an ID card PAD system based on open-access databases.
\end{itemize}

The remainder of the paper is structured as follows: Section \ref{sec:related_word} briefly discusses related works on GANs and ID cards PAD systems. Section \ref{sec:methods} describes the methods used for generating new PA. The data and preprocessing used in the experiments are described in Section \ref{sec:datasets}, while the metrics used to evaluate image quality and PAD predictive performance are presented in Section \ref{sec:metrics}. In Section \ref{sec:experiments}, we detail the applied experimental framework and discuss the results. Finally, we provide a summary of our results in Section \ref{sec:conclusions}.

\section{Related Work} \label{sec:related_word}

In the present section, we introduce the GAN models used to create synthetic presentation attack samples and briefly present fake ID detection systems found in recent literature.

\subsection{Generative Models}

The traditional GAN \cite{goodfellow2014generative} is a pair of neural networks that approaches the data generation task by implicitly modelling the distribution of a given dataset. It is composed of a generator $G$ and a discriminator $D$ network. The generator tries to generate data that is indistinguishable from the real data, whereas the discriminator tries to determine correctly whether a given data sample is real or fake. Both networks are trained simultaneously in a competitive manner, taking the form of a zero-sum game between two players where the objective is to find the Nash equilibrium. 

With traditional GANs, there is scarce control on the output of $G$ since it solely depends on the input noise vector. Conditional GANs were introduced in \cite{mirza2014conditional} that allow for greater control of the output by conditioning $G$ and $D$ on additional data $\mathbf{x}$ while training.

\begin{figure*}[]
    \centering
    \includegraphics[scale=0.60]{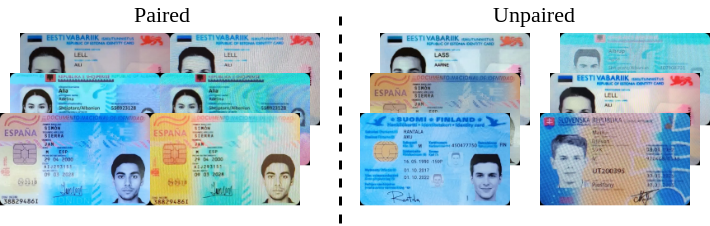}
    \caption{Paired training data (left) consists of pairs of pixel-aligned images. Unpaired data (right) comprises of two sets of images.}
    \label{fig:supervision}
\end{figure*}

Both traditional and conditional GANs are the building blocks of the generative models used in this work. We focused on methods that approach the unimodal image-to-image translation task of finding a mapping $G$ between input $\mathcal{X}$ and output $\mathcal{Y}$ image domains such that $\hat{\mathbf{y}} = G(\mathbf{x})$ is indistinguishable from the images of $\mathcal{Y}$. Four such methods are presented below: pix2pix \cite{isola2017image}, pix2pixHD \cite{wang2018high}, CycleGAN \cite{zhu2017unpaired} and CUT \cite{park2020contrastive}. The first two are supervised methods requiring pixel-aligned data for training, while the last two are unsupervised and trained on unpaired input and output image sets. The difference between paired and unpaired data is shown in Fig. \ref{fig:supervision}.

pix2pix is based directly on the conditional GAN architecture. It uses as input to both networks the concatenation the original input and conditioning images on the channel dimension. In addition to the GAN loss, the authors use a $L1$ loss to enforce correctness at the low frequencies. A U-Net \cite{ronneberger2015unet} architecture is used as the generator, while a novel ``PatchGAN'' architecture is used as the discriminator that classifies patches as real or fake and aggregates the results.

pix2pixHD aims to improve upon pix2pix for high-resolution image generation. The authors introduce a novel coarse-to-fine generator and a multi-scale discriminator architecture, as well as an improved adversarial loss based on matching the discriminator features at different layers. Furthermore, pix2pixHD allows conditioning with instance boundary maps and semantic label maps to improve the rendering of object boundaries.

For certain tasks, obtaining image pairs is a relatively simple process, such as for superresolution, colourization and inpainting. However, for a number of other tasks such as translating photos to landscape paintings it can be a prohibitive endeavor. This motivated the search for methods that could accomplish domain translation between unpaired sets of images.

CycleGAN achieves unpaired image translation by using two GANs and enforcing a cycle consistency loss between the generators. That is, given the generator $G$ from $\mathcal{X}$ to $\mathcal{Y}$ and the generator $F$ from $\mathcal{Y}$ to $\mathcal{X}$, the authors add to the GAN losses the following in Equation \eqref{eq:1}:
\begin{multline}
\label{eq:1}
    \mathcal{L}_{\text{cyc}}(G, F) = \mathbb{E}_{\mathbf{x}}[|| F(G(\mathbf{x})) - \mathbf{x} ||_{1}] \\ + \mathbb{E}_{\mathbf{y}}[|| G(F(\mathbf{y})) - \mathbf{y} ||_{1}]
\end{multline}
The cycle consistency condition, although an effective strategy to approach the unpaired translation problem, tends to force $G$ into generating samples that contain all the necessary information in order to translate back to the input image, which leads to unsatisfactory results if significant visual changes are expected. 

CUT uses a patch-wise contrastive loss \cite{oord2019representation} to maintain content correspondence between input and output images. The loss enforces similarity between corresponding patches of the input and generated images while enforcing dissimilarity with negative patches from the input image. The authors propose to use an encoder-decoder architecture for the generator, where the contrastive losses are computed on patches of features extracted from the encoder.

\subsection{Fake ID detection}

The widespread use of smartphones has prompted the development of novel remote authentication systems embedded in applications that require the input of biometric data such as fingerprints, faces, iris, and selfies. Additionally, many services require digital photographs of ID cards, which are often captured with smartphones, as part of their digital onboarding process. Methods found in the literature that tackle the problem of remotely detecting fake ID cards from digital photographs are presented below.

\citet{berenguel2017counterfeit} developed a novel application to detect ID documents that have been forged by a scan-printing operation. Their application allows the capture of Spanish ID documents using a mobile device and the assessment of their validity. The counterfeit detection module performs texture descriptor extraction, principal component analysis and feature pooling to classify regions of interest with linear Support Vector Machines (SVM). The final decision of labelling a document as genuine or counterfeit is performed by a naïve Bayes classifier. Additionally, \citet{berenguel2019recurrent} proposed a counterfeit document detector that uses a recurrent comparator architecture with attention models to spot the differences between a genuine and a reference image. The authors applied the detector to datasets of Spanish ID documents and banknotes. The system searches for the lack of resolution due to a scanning-printing operation by iteratively centring the attention on different positions of the security background textures and computing the differences.

\citet{gonzalez2021hybrid} presented a two-stage method for detecting tampered ID cards, which was trained and evaluated on a database of real Chilean national ID cards. The proposed method uses a pre-trained MobileNet \cite{howard2017mobilenets} to detect borders in the photo ID zone caused by composite tampering, while a second lightweight CNN, termed ``BasicNet'', was trained from scratch to detect the physical source of the document. 

The DLC-2021 dataset \citet{dlc2021} presented and defined three detection tasks: 1) screen recapture detection, where centre crops of documents from the original frames are classified as bona fide or screen recapture presentation, 2) unlaminated colour copy detection, where the network classifies scaled down images grey images as print presentation of bona fide, and 3) grey copy detection, where the classification is performed on projective undistorted document images. The authors train variations of the ResNet-50 \cite{he2016deep} architecture on each task and report the results for future reference.

\citet{mudgalgundurao2022pixelwise} proposed a pixel-wise supervision methodology which is used, along with a binary classification objective, to train presentation attack detectors on an in-house database of German ID cards and residence permits. The proposed system uses a simplified DenseNet \cite{huang2017densely} architecture, which the authors compare against baseline face PAD approaches.

\citet{chen2022domain} employed a scheme based on Siamese networks for document recapture detection. The network is trained on triplets of patches extracted from bona fide, recaptured, and reference documents. A custom ``forensics loss'' is used to attract genuine and reference representations while repelling recaptured and reference representations. The authenticity of a questioned document is evaluated using the distance metrics from three triplets. The authors created a database of synthetic university student ID cards to test their system.

\citet{benalcazar2023synthetic} explored the effectiveness of computer vision algorithms and generative models for the purpose of data augmentation while training fraud detection networks. The authors propose populating templates with synthetic data to create additional bona fide presentations, as well as training a StyleGAN-ADA \cite{viazovetskyi2020stylegan2} network to generate synthetic bona fide samples from scratch. For creating attack presentations, they use, in addition to the latter network, a texture transfer method based on adding artificial textures to bona fide presentations and a CycleGAN \cite{zhu2017unpaired} model to translate between bona fide and attack domains. Various MobileNetV2 \cite{sandler2018mobilenetv2} models were trained on different combinations of real and synthetic Chilean national ID card presentations to assess the quality of the generated images. The authors report a negligible performance loss when supplementing databases with synthetic images. 

\citet{magee2023investigation} explored the potential application of the Meijering filter \cite{meijering2004design} to the domain of recaptured identity document detection. The authors create a new dataset of recaptured images based on the publicly available BID \cite{alysson2020bid} dataset and use it to train an SVM classifier on the raw histogram data obtained by using the filter. Although their system does not compare well with approaches that utilize neural networks, it remains an attractive alternative due to being transparent and explainable.

Most of the aforementioned studies train and test their proposed systems on private datasets using presentations of bona fide ID cards obtained from  Gubernamental entities, company services, and banks in order to prevent fraud. As such, it is difficult to scrutinize and improve upon these systems since the data can not be distributed publicly due to privacy concerns. In light of these challenges, some studies, though few in number, have created and published datasets composed of synthetic ID cards generated from templates, as seen in Table \ref{tab:dataset_characteristics_studies}. However, they have limited commercial applicability because of the reduced number of subjects for each bona fide and attack image. These efforts are crucial for the effective public benchmarking of novel PAD systems. 

\begin{table}[t!]
\centering
\caption{ID card dataset availability in the SOTA}
\label{tab:dataset_characteristics_studies}
\begin{tabularx}{\linewidth}{ l X c }
\hline
\hline
\textbf{Author} & \textbf{ID card type} & \textbf{Open-Access} \\ \hline
\citet{berenguel2017counterfeit} & Spanish national & \ding{55} \\
\citet{berenguel2019recurrent} & Spanish national & \ding{55} \\
\citet{gonzalez2021hybrid} & Chilean national & \ding{55} \\
\citet{dlc2021} & Various national & \ding{51} \\
\citet{mudgalgundurao2022pixelwise} & German national and residence permits & \ding{55} \\
\citet{chen2022domain} & University student & \ding{51} \\
\citet{benalcazar2023synthetic} & Chilean national & \ding{55} \\
\citet{magee2023investigation} & Brazilian national & \ding{55} \\
\hline
\hline
\end{tabularx}
\end{table}

\section{Methods} \label{sec:methods}

This section details the implementation of the GAN-based models used to generate additional presentation attack samples and of the PAD system used for both the screen and print tasks.

The proposed generative methods are designed to create synthetic PA of full ID cards. The composition of the datasets used to train each model is detailed in Section \ref{sec:datasets}. The ID cards under study have widths of 85-86 mm and heights of 53-55 mm, while the employed image generator requires the input width and height to be a multiple of $8$; thus an image size of $448\times728$ was chosen after preprocessing. The training of the generative models is performed on crops of these images, whereas the full images are used for inference.

\subsection{Image Generation from GANs}

The GAN-based image-to-image translation methods used in this work to generate new samples of presentation attacks from bona fide presentations are pix2pix, pix2pixHD, CycleGAN and CUT. The first two are trained on paired data, while the remaining two simply require unpaired sets of images, one set per domain. 

For the print task, the methods learn to transfer the visual characteristics of printed documents, such as the paper texture and fine-grain elements left by printers and ink. On the other hand, the pixel grid texture, spatial aliasing, and colour distortions indicative of screen displays are expected to be learned and faithfully transferred for the screening task.

An automatic procedure to generate pixel-aligned paired training data for pix2pix and pix2pixHD was implemented. Firstly, each bona fide image is paired randomly with a presentation attack image of the same subject. Next, the ORB algorithm \cite{rublee2011orb} is used on each image to detect keypoints and extract binary local invariant features. Then, the Hamming distance between the features of one image and the features of the other is computed, and the best matches are found using an iterative algorithm. Finally, the homography matrix is estimated from the comparisons and is applied to align the presentation attack to the bona fide presentation.

For all the methods, we adopt the generator architecture from \cite{johnson2016perceptual} with 9 residual blocks. Additionally, the $70\times70$ PatchGAN \cite{isola2017image} architecture with 4 convolutional layers was used for the discriminator. We used three PatchGAN models for the multi-scale discriminator of pix2pixHD.

Training for each method and task was performed for 200 epochs with a batch size of 1 on random crops of size $224\times224\times3$. Adam \cite{kingma2015adam} was used as the optimizer with an initial learning rate of $2e-4$ and $\beta_{1}=0.5, \beta_{2} = 0.999$. The learning rate for pix2pix was maintained fixed throughout training, while for the other methods, it declined linearly to zero after the 100th epoch. During inference, bona fide presentations of size $448\times728\times3$ are fed to the generators to produce synthetic PA of the same size. The execution of training and inference was performed on a server with 32 CPU cores, 236 GB of RAM and a GPU of 40GB.

\subsection{Fraud-Detection Networks}

The same network architecture was used for both the screen and print tasks. Following \cite{benalcazar2023synthetic}, we used as the backbone a MobileNetV2 \cite{sandler2018mobilenetv2} pre-trained on ImageNet. The input to the networks is the $448\times448\times3$ center crop of each image normalized with ImageNet mean and variance. The weights of the backbone are frozen during training. The output of the backbone is fed to a dropout layer \cite{srivastava2014dropout} with $p=0.2$, and the result is in turn fed to a final linear layer.

We use a batch size of 128 and train for 100 epochs. The weights are optimized using AdamW \cite{loshchilov2018decoupled} with a constant learning rate of $5e-4$. Training and inference were performed on the same server as the generative models, with 32 CPU cores, 236 GB of RAM, and a GPU of 40GB.

\section{Datasets} \label{sec:datasets}

This section describes the datasets used in this work. Most of them present an important variability in light, illumination, background, orientation and others. It is essential to highlight that estimating the capture quality of ID cards is still an open problem \cite{Schulz-DIQA}.

Specifically, the source datasets are presented, and the print and screen tasks are formally defined (See Table \ref{tab:doc_types}. We use approximately 48,350 images derived from open-source datasets for our study. Furthermore, around 21,700 images of synthetic PA were generated for augmenting PAD model training sets.

Two experiments are defined: 
\begin{itemize}
    \item Experiment 1: the ``print'' task, where the PAD systems are meant to distinguish between bona fide and coloured print attack presentations.

    \item Experiment 2: the ``screen'' task, where bona fide and screen attack presentations are differentiated. The details of how the dataset for each task is constructed are provided below.
\end{itemize}
 The list of images used to replicate and compare this proposal will be available for research purposes (upon paper acceptance).

\subsection{Dataset organisation}

The MIDV-2020 and DLC-2021 datasets were used for this work\footnote{MIDV-2020 is available to download from ftp://smartengines.com/midv-2020/dataset/. DLC-2021 is available in three parts from https://zenodo.org/records/7467028, https://zenodo.org/records/7467004 and https://zenodo.org/records/7467000.}.
Both are successors of MIDV-500 and consist of short video clips of fake documents presented in different lighting and background situations. 

In MIDV-2020, the physical, bona fide documents were captured vertically in a resolution of $2,160\times3,840\times3$ pixels with 60 frames per second using a Samsung S10 or an Apple iPhone XR. In DLC-2021, physical and printed documents were captured, as well as screen presentations of the templates, where two office desktops and two notebook LCD monitors were used. The capturing was done with the same devices in two different frame resolutions ($1,080\times1,920\times3$ and $2,160\times3,840\times3$) and two different frame rates (30, 60 frames per second). In both datasets, the frames of the clips were extracted, and the authors annotated the position of the document.


The splits defined for the print and screen tasks were defined on the subject level. Each split contains at least two subjects for each type of document. Furthermore, bona fide representations from DLC-2021 of subjects with PA are included in the corresponding split. Additional subjects are then added from MIDV-2021 to ensure an approximately equal number of presentations for each class. In what follows, we specify which subjects from DLC-2021 and which from MIDV-2021 are included in each split for each task.

\subsubsection{Print task}

From DLC-2021, subjects 04-07 were used for the training data split, subjects 02 and 03 for validation data and subjects 00 and 01 for test data. From MIDV-2021, subjects 21-27 were used for the training data, subjects 32-38 for validation except for Albanian subject 35, and subjects 39-43 for test data.

\subsubsection{Screen task}

Subjects from DLC-2021 included in the training set are Albanian subjects 04 and 05, Spanish subjects 04, 05 and 06, Estonian subjects 04, 06 and 07, Finnish subjects 04, 05 and 07 and Slovakian subjects 04-07. The validation set contains DLC-2021 subjects 02 and 03, while the test set contains subjects 00 and 01. The training set contains no additional subjects. 

From MIDV-2021, the validation includes Albanian subjects 17-22, Spanish, Estonian and Finnish subjects 18-23 and Slovakian subjects 19-24, while the additional subjects of the test set are Albanian subjects 09-15, Spanish, Estonian and Finnish subjects 10-16 and Slovakian subjects 11-17. 


In both tasks, bona fide presentations come from both MIDV-2020 and DLC-2021, whereas attack presentations originate exclusively from DLC-2021. Moreover, only the data corresponding to ID cards, shown in Table \ref{tab:doc_types}, are considered, and images where the documents lie partially outside the frame are discarded. 

\begin{table}[t!]
\centering
\caption{Types of documents used in the experiments}
\label{tab:doc_types}
\begin{tabular}{ c l l }
\hline
\hline
\textbf{Document type code} & \textbf{Description} & \textbf{PRADO code} \\ \hline
alb\_id & Albanian ID document & ALB-BO-01001 \\
esp\_id & Spanish ID document & ESP-BO-03001 \\
est\_id & Estonian ID document & EST-BO-03001 \\
fin\_id & Finnish ID document & FIN-BO-06001 \\
svk\_id & Slovakian ID document & SVK-BO-05001 \\
\hline
\hline
\end{tabular}
\end{table}

The splits for each task are done on the subject level. At least two subjects must be present in each split and class. Furthermore, for the purpose of a fair comparison, we force the number of samples per class to be the same for each split. 

Table \ref{tab:datasets_stats}, contains the number of images per split and class for the print and screen tasks. Approximately 13,500 images are assigned to the training set of the print task, followed by 6,500 images in the validation set and 6,700 in the test set. The train set of the screen task has 7,960 images, the validation set has 6,460 images, and the test set has 7,230 images. 

\begin{figure*}[]
    \centering
    \includegraphics[scale=0.50]{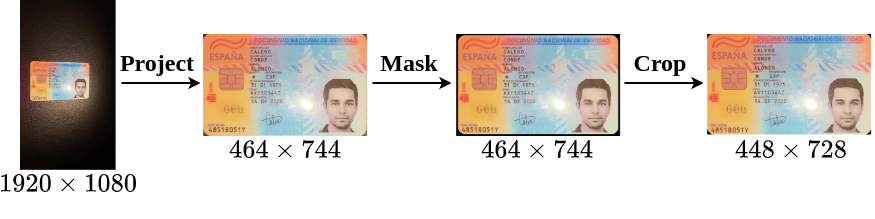}
    \caption{Preprocessing steps applied to the raw frame data.}
    \label{fig:preprocessing}
\end{figure*}

Once the splits are created, the raw frames are preprocessed offline in three steps: firstly, the documents are projected to a $464\times744\times3$ rectangle via a perspective transformation using an estimated homography matrix; secondly, a portion of the background is masked out, and lastly, a centre crop of $448\times728\times3$ is applied. This process is illustrated in Fig. \ref{fig:preprocessing}. Although the annotations were used to perform the projection in the first step, automatic segmentation of the document is possible with networks such as the one proposed by \citet{lara2021towards}.

\begin{table}[t!]
\caption{Number of images per partition and class for each task}
\label{tab:datasets_stats}
\centering
\begin{tabular}{ l l r r r }
\hline
\hline
\textbf{Exp} & \textbf{Class} & \textbf{Train} & \textbf{Validation} & \textbf{Test} \\ \hline
\multirow{2}{*}{Exp1: Print} & Bona fide & 6,782 & 3,285 & 3,317  \\
& Print & 6,720 & 3,212 & 3,386 \\ \hline
\multirow{2}{*}{Exp2: Screen} & Bona fide & 4,066 & 3,224 & 3,633  \\
& Screen & 3,891 & 3,239 & 3,596 \\
\hline
\hline
\end{tabular}
\end{table}

\subsection{Training sets of real and synthetic images}

The training set of each task is split into two disjoint sets, denoted by $\mathcal{T}_{A}$ and $\mathcal{T}_{B}$, in order to adequately assess the impact of using synthetic data. We denote by $\mathcal{T}^{b}_{B}$ the set of bona fide presentations of $\mathcal{T}_{B}$, which is used to generate the set of synthetic PA $\mathcal{T}^{s}_{B}$ for each method. Additionally, to reduce bias, we apply a mask to remove background information from generated images. The sizes of these datasets are shown in Table \ref{tab:training_sets}. Section \ref{sec:experiments} describes how these sets are combined to create the training sets for the PAD systems so that the effect of synthetic data can be quantified. 

\begin{table}[t!]
\caption{Sets of bona fide and synthetic images used to create the experiment training sets}
\label{tab:training_sets}
\centering
\begin{tabular}{ l l r r r r }
\hline
\hline
\textbf{Task} & \textbf{Class} & $\mathcal{T}_{A}$ & $\mathcal{T}_{B}$ & $\mathcal{T}^{b}_{B}$ & $\mathcal{T}^{s}_{B}$ \\ \hline
\multirow{2}{*}{Print} & Bona fide & 3,391 & 3,391 & 3,391 & 0 \\
 & Print & 3,360 & 3,360 & 0  & 3,391 \\ \hline
\multirow{2}{*}{Screen} & Bona fide & 2,033 & 2,033 & 2,033 & 0 \\
 & Screen & 1,945 & 1,946 & 0 & 2,033\\ 
\hline
\hline
\end{tabular}
\end{table}

\section{Metrics} \label{sec:metrics}

This section describes the metrics used, on the one hand, to evaluate the quality of generated images and, on the other, to assess the predictive performance of PAD systems.

\subsection{Quality of Generated Images}

Effective research concerning generative models relies upon metrics that can meaningfully assess the quality of generated images. In recent years many quantitative methods for computing the quality and diversity of synthetic data have been developed hand in hand with novel generative architectures, such as the Inception Distance \cite{salimans2016advances}, the Fr{é}chet Inception Distance (FID) \cite{heusel2017advances} and the Kernel Inception Distance \cite{binkowski2018demystifying}. The FID was selected for use in order to compare the results with the SOTA.

The FID is a similarity metric between two probability distributions that in practice, is often used to assess the similarity between two sets of images $X$ and $Y$. To calculate the FID, firstly, the 2,048 dimensional feature vector (embedding) for each image is obtained by processing the image with a pre-trained Inception-v3 \cite{szegedy2016rethinking} network and keeping the features from the pool3 layer. Secondly, the per image set mean vectors $\mu_{X}, \mu_{Y}$ and covariance matrices $\Sigma_{X}, \Sigma_{Y}$ are calculated, and finally, the distributions are compared using the following Equation \eqref{eq:fid}:
\begin{multline}
\label{eq:fid}
    \text{FID} = ||\mu_{X} - \mu_{Y}||_{2}^{2} \\ + \mathrm{Tr}\left( \Sigma_{X} + \Sigma_{Y} - 2 (\Sigma_{X} \cdot \Sigma_{Y})^{1/2} \right)
\end{multline}
where $||\cdot||^{2}_{2}$ is the squared $L^2$ distance and $\mathrm{Tr}(\cdot)$ is the trace function.

\subsection{Detection Performance Evaluation}

The methodologies for evaluating the detection performance of biometric PAD algorithms are standardized by the ISO/IEC 30107-3\footnote{\url{https://www.iso.org/standard/79520.html}} standard. The metrics used by this study are Attack Presentation Classification Error Rate (APCER), Bona fide Presentation Classification Error Rate (BPCER), BPCER\textsubscript{AP} and Equal Error Rate (EER). These metrics are aggregates of comparisons between the ground truth label $y \in \{0, 1, \ldots, J\}$ and the prediction $\hat{y}(\tau) \in \{0, 1, \ldots, J\}$ for a given operating point $\tau \in [0, 1]$, where $y = 0$ indicates a bona fide representation and $y = j$ with $j \geq 1$ is a presentation of the $j$th attack type.

To compute the APCER, firstly, the percentage of attack presentations incorrectly classified as bona fide is calculated for each presentation attack instrument as is shown in Equation \eqref{eq:APCER}:
\begin{equation}
\label{eq:APCER}
    \text{APCER}_{j}(\tau) =  \frac{100}{\sum_{i=1}^{N}[y_{i} = j]} \sum_{i=1}^{N} [y_{i} = j] [\hat{y}_{i}(\tau) = 0]
\end{equation}
where $[\cdot]$ is the Iverson bracket and $N$ is the total number of presentations. Lastly, the maximum of these values, the worst-case scenario, is considered:
\begin{equation}
\text{APCER}(\tau) = \underset{j}{\max} \; \text{APCER}_{j}(\tau)
\end{equation}
On the other hand, the BPCER is the percentage of bona fide presentations that have not been classified as is shown in Equation \eqref{eq:BPCER}:
\begin{equation}
\label{eq:BPCER}
    \text{BPCER}(\tau) = \dfrac{100}{\sum_{i=1}^{N}[y_{i} = 0]}\sum_{i=1}^{N} [y_{i} = 0][\hat{y}_{i}(\tau) \neq 0]
\end{equation}
The remaining two metrics analyze the system performance on specific operating points. The BPCER\textsubscript{AP} is the BPCER value when the APCER is fixed at $100/\text{AP}$. In this work we evaluate BPCER\textsubscript{10}, BPCER\textsubscript{20} and BPCER\textsubscript{100}, which correspond to APCER values of 10\%, 5\% and 1\% respectively. The EER is the operating point where $\text{APCER} = \text{BPCER}$; however, the classification rate is often reported instead. In practice, there may not exist an operating point that satisfies the previous condition, thus, a reasonable interpolated value is often used. 

The aforementioned metrics can be represented using Detection Error Tradeoff (DET) curves \cite{martin2011det}, which are also used by this study. They represent the APCER on the $X$ axis and the BPCER on the $Y$ axis and use a normal deviate scale for both axes, which spreads out the plot and facilitates the visual comparison of different systems.

\section{Experiments and Results} \label{sec:experiments}

This section describes the experiments performed on the data described in Section \ref{sec:datasets} and reports the results using the metrics shown in Section \ref{sec:metrics}. First, we show how generative methods perform in terms of the visual similarity of generated samples with real data. Then, for each task, we analyze the effect on PAD predictive performance of adding synthetic samples to the training dataset instead of bona fide samples, which, in practice, are harder to obtain.

\subsection{Synthetic Image Quality Evaluation} \label{sec:quality_experiments}

The aim of this experiment is to assess the quality of synthesized PA by comparing them to sets of real PA. The comparison between the two sets of images is done with the FID metric. Each generative system was trained on $\mathcal{T}_{A}$ and applied on $\mathcal{T}_{B}^{b}$ to obtain the set of synthesized presentations $\mathcal{T}_{B}^{s}$. Then, the FID of $\mathcal{T}_{B}^{s}$ with the test presentation attack images is computed. This allows for the comparison of generative methods in terms of the visual quality of the generated samples.
Additionally, the obtained FID values are compared to our baseline FID values obtained from the validation set in order to compare the difference between synthetic images and PA (screen, print).

Table \ref{tab:official_fid}, shows the results in terms of the FID scores. The second column, which reports the best results, shows the baseline FID scores. With a FID score of 43.24 for the print task and of 58.22 for the screen task, these values represent the ideal performance for a generative model. The best performing FID computed on synthetic data is shown in bold for each task, where CycleGAN obtains the best results for both tasks with FID scores of 55.73 and 68.77 for the print and screen task respectively.

\begin{table}[t]
\caption{FID scores computed between synthetic fake images with proposed generation methods and test data}
\label{tab:official_fid}
\centering
\begin{tabular}{l r r r r r}
\hline
\hline
\textbf{Task} & \textbf{Validation} & \textbf{CycleGAN} & \textbf{CUT} & \textbf{pix2pix} & \textbf{pix2pixHD} \\ \hline
Print & 43.24 &  \B 55.73 & 56.69 & 60.00 & 65.02  \\
Screen & 58.22 & \B 68.77 & 77.39 & 78.92 & 76.16  \\
\hline
\hline
\end{tabular}
\end{table}

In regards to synthesized print attacks, CycleGAN is closely followed by CUT, while pix2pix and pix2pixHD are the worst-performing methods with FID scores above 60. On the other hand, in terms of synthesized screen attacks, the next best method is pix2pixHD, closely followed by CUT and lastly pix2pix. 

\begin{figure*}[]
    \centering
    \includegraphics[width=\textwidth]{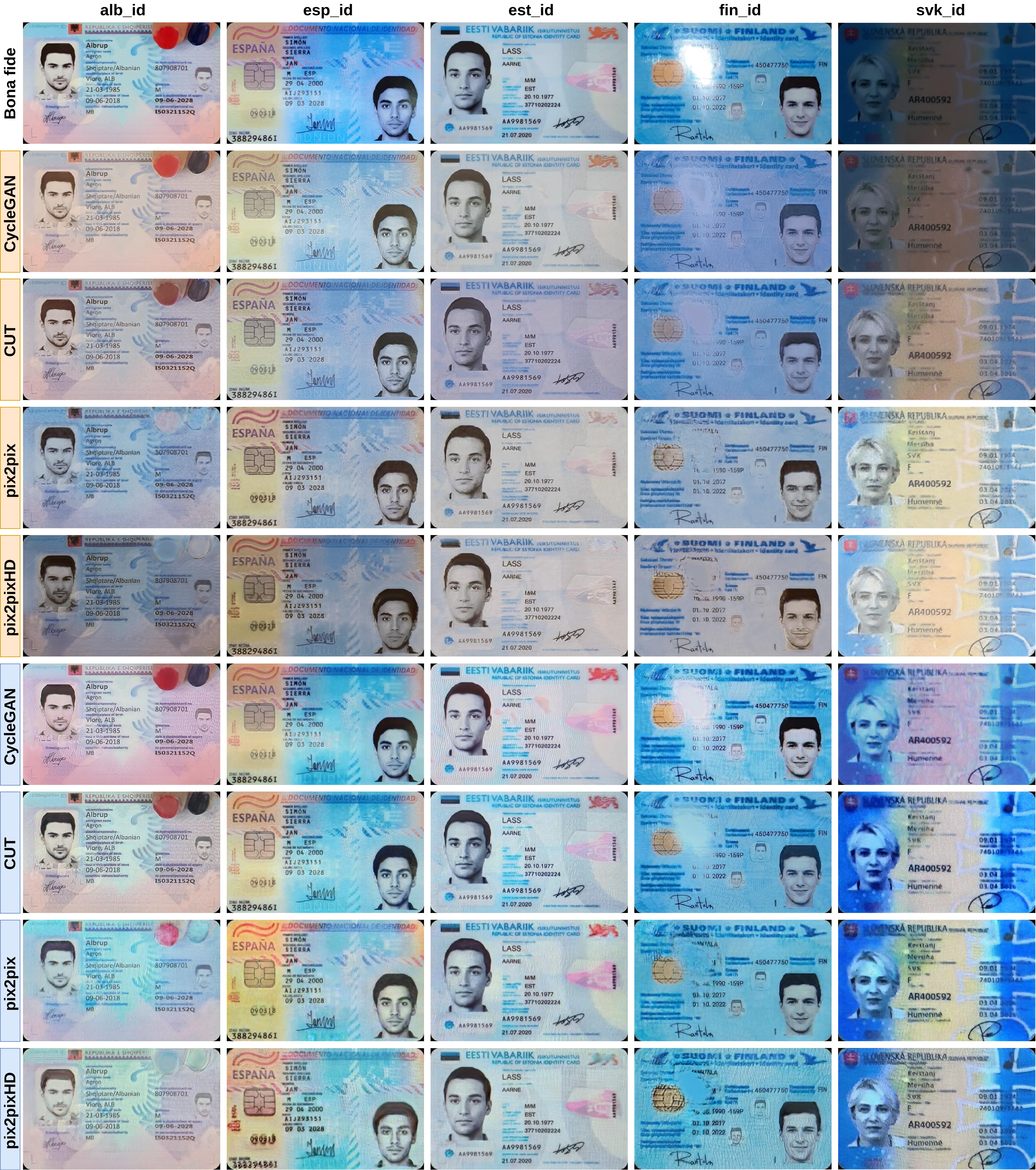}
    \caption{Examples of bona fide and synthetic images generated with the proposed methods. The first row contains bona fide samples of each type of ID card, and the following rows contain the output produced by each method. The method names with orange backgrounds were trained on the print task, while the ones with blue backgrounds were trained on the screen task.}
    \label{fig:examples}
\end{figure*}

For both types of attack, we observe that unsupervised methods (CycleGAN \& CUT) perform on par or better than supervised methods. We hypothesize that this is due in part to deficiencies in the automatic image alignment process to produce paired training data, which results in noisier images. Additionally, we suspect that the unnaturalness of the reconstructions of external elements in the source image, such as fingers or reflected light, could contribute to a higher FID score. These effects can be observed in Fig. \ref{fig:examples}, in which bona fide presentations of each document type, as well as the corresponding synthetic PA generated with each method, are presented.

The examples generated with CycleGAN shown in Fig. \ref{fig:examples} preserve better the content of the bona fide presentation compared with samples of the same attack type generated with other methods. This is likely a result of the increased capacity of the cycle consistency loss for conserving information between translations. Moreover, in addition, the aforementioned problems with the supervised data, the increased complexity of the discriminator, and the low resolution and diversity of the training data are likely causes of the increased noise and artefacts observed in the samples obtained with pix2pixHD.   

\begin{figure*}
\centering
\begin{subfigure}{0.3\textwidth}
    \includegraphics[width=\textwidth]{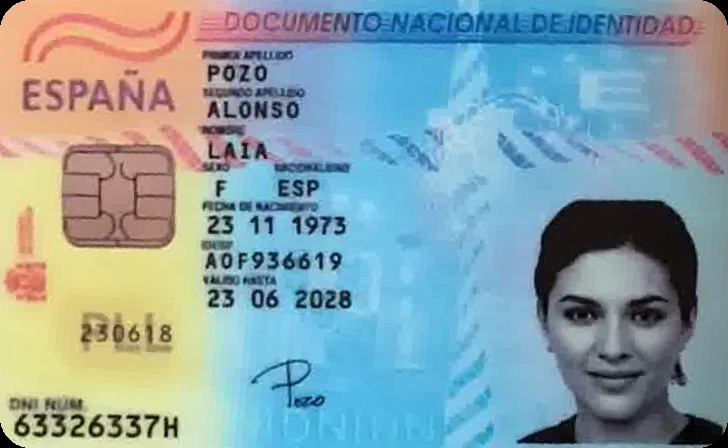}
    \caption{esp\_id bona fide}
\end{subfigure}
\hfill
\begin{subfigure}{0.3\textwidth}
    \includegraphics[width=\textwidth]{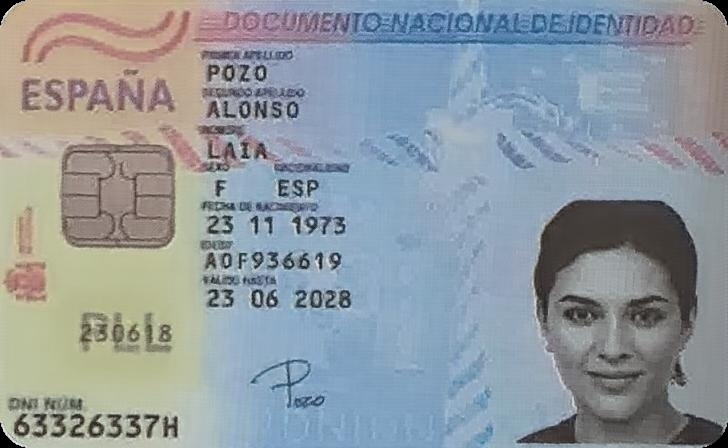}
    \caption{esp\_id generated print}
\end{subfigure}
\hfill
\begin{subfigure}{0.3\textwidth}
    \includegraphics[width=\textwidth]{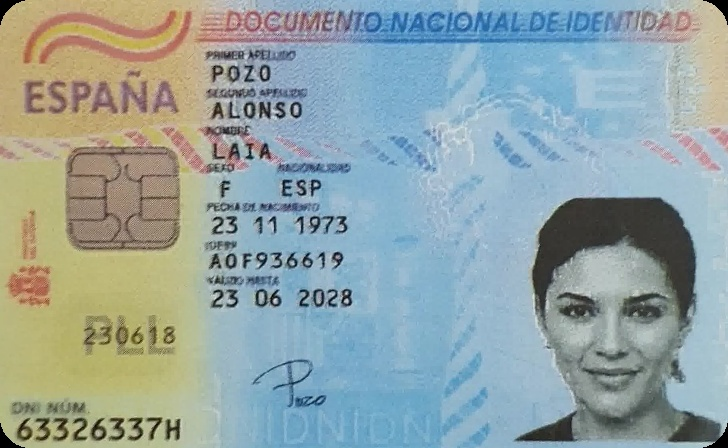}
    \caption{esp\_id print}
\end{subfigure}
\par\bigskip 
\begin{subfigure}{0.3\textwidth}
    \includegraphics[width=\textwidth]{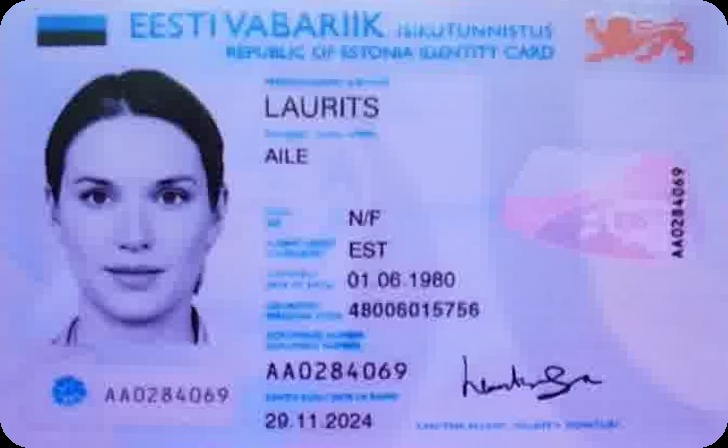}
    \caption{est\_id bona fide}
\end{subfigure}
\hfill
\begin{subfigure}{0.3\textwidth}
    \includegraphics[width=\textwidth]{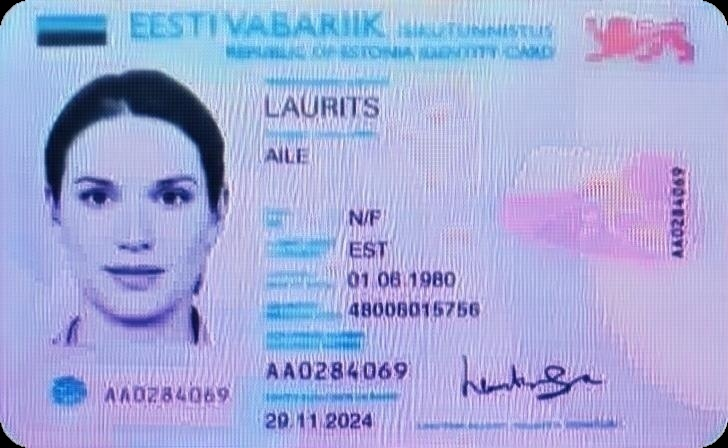}
    \caption{est\_id generated screen}
\end{subfigure}
\hfill
\begin{subfigure}{0.3\textwidth}
    \includegraphics[width=\textwidth]{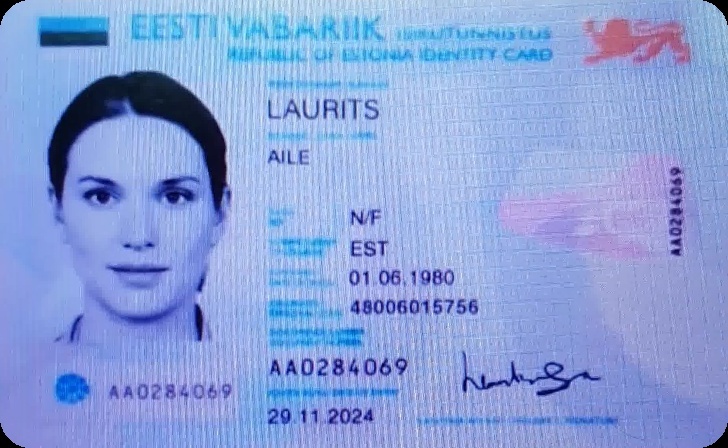}
    \caption{est\_id screen}
\end{subfigure}
        
\caption{Examples of ID cards. (a) to (c) showcase examples of bona fide, generated print presentations and bona fide print presentations of Spanish ID cards. (d) to (f) showcase examples of bona fide, generated screen presentations and bona fide screen presentations of Estonian ID cards. (b) and (e) were generated with CycleGAN.}
\label{fig:dlc2021-matrix}
\end{figure*}

Additionally, from Fig. \ref{fig:dlc2021-matrix}, it can be seen that there are noticeable differences in colour and brightness between the synthetic print presentations of each bona fide document. This may be due to a number of factors, such as the variability of lighting conditions, the presence of different document types in the dataset or the differences between the methods of preserving input colour information. However, the colour saturation of the generated samples appears to be lower than that of the corresponding bona fide presentation, which is expected when printing on matte or uncoated papers. The differences in colour and brightness between synthesized screen presentations seem to be less pronounced for certain documents, which can be attributed to the reduced variability of these aspects in screen displays. Moreover, some synthetic samples present a grid-like texture, which shows that the generative models have successfully learned to transfer this feature of screen displays. On the other hand, there is an absence of sophisticated moir{é} patterns in the generated samples, which is likely due to the small number of training samples with such patterns, as well as the distortion of the original patterns due to the projection of the document segments in the preprocessing stage.

\subsection{Experiment 1: PAD Performance on print task}

For the print task experiments, the MobileNetV2 networks are binary image classifiers that detect whether the input presentation is bona fide or a print attack. In total, 6 networks were trained. 
The first network was trained using only $\mathcal{T}_{A}$ (6,751 images) in order to gauge the effect of adding more data. The second network was trained on $\mathcal{T}_{A} \cup \mathcal{T}_{B}$ (13,502 images) and represents the model trained with the complete set of real data. 
The remaining networks were trained on $\mathcal{T}_{A} \cup \mathcal{T}^{b}_{B} \cup \mathcal{T}^{s}_{B}$ (13,533 images), where $\mathcal{T}^{s}_{B}$ is different for each method, and represent the cases where synthetic data is used. The validation set of Table \ref{tab:datasets_stats} was used to determine the best checkpoint, and the PAD metrics were calculated on the test set.

\begin{figure}[ht]
    \centering
    \includegraphics[width=0.4\textwidth]{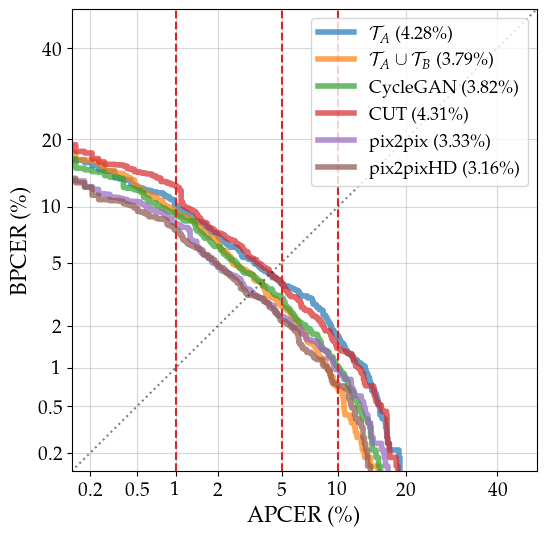}
    \caption{Detection Error Trade-off curves of networks trained on \textbf{print} task data. The EER is shown in parentheses for each scenario.  Red dot lines represent three operational points BPCER\textsubscript{10} and BPCER\textsubscript{20} and BPCER\textsubscript{100}, respectively.}
    \label{fig:print_det_curves}
\end{figure}

The DET curves obtained from predictions of the print task test set of all networks are displayed in Fig. \ref{fig:print_det_curves}, where the EER values for each curve are also reported. Interestingly, the best performance is obtained with synthetic PA generated with pix2pixHD, with an EER of 3.16\%. With pix2pix data, we observe a slight drop in performance with an EER of 3.33\%, but still better than using real PA were the value of 3.79\% was observed. With CycleGAN data, we obtain an EER of 3.82\%, which is comparable to using real data. CUT produced the worst performing data with an EER of 4.31\%, which is slightly above the 4.28\% obtained by only using $\mathcal{T}_{A}$.

Table \ref{tab:metrics_print} contains the BPCER\textsubscript{10}, BPCER\textsubscript{20} and BPCER\textsubscript{100} operational points of each experiment. We observe a similar trend as reported with the EER values, except for BPCER\textsubscript{10} where pix2pixHD performs on par with real data with a value of 0.72\% and all cases perform better than training only with $\mathcal{T}_{A}$ where a value of 1.90\% was observed.

Given the previous observations, it can be said that data synthesized using supervised generative models produced better-performing PAD models than data generated with unsupervised methods. However, we observed in Section \ref{sec:quality_experiments} that supervised methods produced the data most dissimilar to real data. Hence, when synthetic print attacks are involved, the FID score correlates positively with PAD predictive performance.

\begin{table}[t!]
\caption{Results for PAD models trained on the print task. Values expressed in \%}
\label{tab:metrics_print}
\centering
\resizebox{0.48\textwidth}{!}{%
\begin{tabular}{ l r r r r r r }
\hline
\hline
 & $\mathcal{T}_{A}$ & $\mathcal{T}_{A} \cup \mathcal{T}_{B}$ & \textbf{CycleGAN} & \textbf{CUT} & \textbf{pix2pix} & \textbf{pix2pixHD} \\ \hline
  EER & 4.28 & 3.79 & 3.82 & 4.31 & 3.33 & \B 3.16 \\
 $\text{BPCER}_{10}$  & 1.69 & 0.72 & 1.03 & 1.39 & 0.96 & \B 0.72 \\
 $\text{BPCER}_{20}$ & 3.80 & 2.71 & 3.04 & 3.83 & 2.32 & \B 2.17  \\
 $\text{BPCER}_{100}$ & 10.16 & 9.35 & 9.20 & 12.63 & 8.20 & \B 7.60 \\
\hline
\hline
\end{tabular}
}

\end{table}

\subsection{Experiment 2: PAD Performance on-screen task}

The screen task models are trained to distinguish between bona fide and screen presentations of ID cards. We configured the experiments in a similar manner to those of the print task, where 6 networks are trained using different combinations of the screen task datasets described in Table \ref{tab:training_sets}. 
The first evaluation uses $\mathcal{T}_{A}$ (3,978 images) as a training set, while the second uses the complete training set $\mathcal{T}_{A} \cup \mathcal{T}_{B}$ (7,957 images) comprised only of real data. The remaining experiments combine the first half of the training set with the bona fide images of the second half and the synthetic images generated from said bona fide images, that is $\mathcal{T}_{A} \cup \mathcal{T}^{b}_{B} \cup \mathcal{T}^{s}_{B}$ (8,044 images). After training, the best-scoring network checkpoint on the validation set is retained and evaluated on the test set of Table \ref{tab:datasets_stats}.

\begin{figure}[ht]
    \centering
    \includegraphics[width=0.4\textwidth]{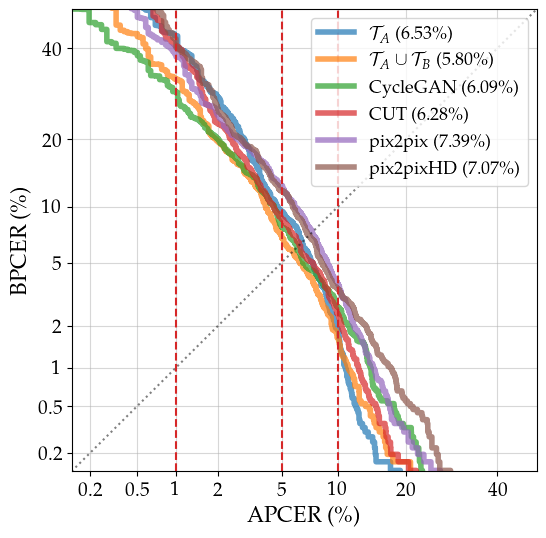}
    \caption{Detection Error Trade-off curves of networks trained on-\textbf{screen} task data. The EER is shown in parentheses for each scenario. Red dot lines represent three operational points BPCER\textsubscript{10} and BPCER\textsubscript{20} and BPCER\textsubscript{100}, respectively.}
    \label{fig:screen_det_curves}
\end{figure}

The DET curve of each screen task experiment is shown in Fig. \ref{fig:screen_det_curves} along with the corresponding EER score. The full training set of real data produces the best model with an EER of 5.80\%. The next best model was trained with CycleGAN data with a score of 6.09\%, followed by the one trained with CUT data with a score of 6.28\%. The aforementioned models have better predictive performance in terms of EER score than the model trained only with $\mathcal{T}_{A}$, where a value of 6.53\% was observed. The worst-performing models were trained on data generated with pix2pixHD and pix2pix, with EER scores of 7.07\% and 7.39\%, respectively.

\begin{table}[t!]
\caption{Results for PAD models trained on the screen task. Values expressed in \%}
\label{tab:metrics_screen}
\centering
\resizebox{0.48\textwidth}{!}{%
\begin{tabular}{ l r r r r r r }
\hline
\hline
 & $\mathcal{T}_{A}$ & $\mathcal{T}_{A} \cup \mathcal{T}_{B}$ & \textbf{CycleGAN} & \textbf{CUT} & \textbf{pix2pix} & \textbf{pix2pixHD} \\ \hline
EER & 6.53 & \B 5.80 & 6.09 & 6.28 & 7.39 & 7.07 \\
$\text{BPCER}_{10}$  & 1.90 & \B 1.62 & 2.67 & 2.39 & 3.72 & 3.47 \\
$\text{BPCER}_{20}$ & 9.50 & \B 7.10 & 7.90 & 8.75 & 12.61 & 11.89 \\
$\text{BPCER}_{100}$ & 43.24 & 32.56 & \B 29.64 & 40.77 & 38.67 & 42.42 \\
\hline
\hline
\end{tabular}
}
\end{table}

A different trend can be observed from the BPCER\textsubscript{AP} values, reported in Table \ref{tab:metrics_screen}. The model trained on $\mathcal{T}_{A} \cup \mathcal{T}_{B}$ obtained the best BPCER\textsubscript{10} and BPCER\textsubscript{20} scores, with values of 1.62\% and 7.10\% respectively, while the model trained with CycleGAN data obtained the best BPCER\textsubscript{100} score with a value of 29.64\%. Furthermore, all models trained with synthetic data obtained worse BPCER\textsubscript{10} scores than the model trained solely on $\mathcal{T}_{A}$ where a value of 1.90\% was observed. BPCER\textsubscript{20} scores show a similar trend as the EER scores. With respect to the BPCER\textsubscript{100} scores, the worst performing model was trained on $\mathcal{T}_{A}$ where a value of 43.24\% was observed, and pix2pixHD provided the worse performing synthetic training data with an observed value of 42.42\%.

The results presented above show that the best-performing synthetic data was provided by unsupervised generative models, with CycleGAN performing better in general than CUT. Moreover, pix2pixHD performed better than pix2pix in three out of four metrics. This follows the same trend as seen with the FID scores seen in Table \ref{tab:official_fid}. We postulate that the noise introduced by the automatic pixel-alignment process hinders the performance of PAD models because they rely on high-level texture details to detect screen presentations adequately. A closer look at the presentations of the last four rows of Fig. \ref{fig:examples} shows how the unsupervised models produce samples with greater detail and finer texture than supervised models. Besides, in some samples generated by the latter models, the alphanumeric information appears distorted, which might further degrade predictive performance.

\subsection{Discussion}

The main focus of this article is to analyze the potential of GAN-based methods for generating effective PA of front-faced, full ID cards on open-access datasets. For this purpose, we first compared the generated data with real data using the FID metric, and then we evaluated their contribution to PAD predictive performance when replacing real data. However, the reported results are not directly comparable to those of SOTA. 
Firstly, most of the papers in the SOTA using a private dataset are not available.
Second, most studies used different types of ID cards (countries) than those used here and with different subject numbers. Third, all cited studies train their respective PAD models using non-projected data. Fourth, some studies use datasets comprised of more than one type of attack for training, as is the case in \citet{benalcazar2023synthetic}. Lastly, some studies, notably \citet{dlc2021}, fail to report predictive performance using ISO/IEC 30107-3 standardized metrics.

\section{Conclusions} \label{sec:conclusions}

In this work, we addressed the problem of open-access PA data scarcity by proposing methods that use GAN-based generative models to create synthetic samples. Additionally, we studied whether the generated data are an effective substitute for real data for training PAD models. For this purpose, we leveraged two open-source datasets containing ID card presentations of fake subjects. 

We defined two experiments based on these datasets: the ``print'' task to distinguish between bona fide and print presentations and the ``screen'' task to differentiate between bona fide and screen presentations. We trained a total of $12$ MobileNetV2 networks, 6 for each task, using different combinations of datasets comprised of real and generated data in order to adequately assess the impact of adding more training data and replacing real PA with synthesized ones. 

The results vary greatly from one task to the other. Regarding the print task, data generated with the supervised generative models proved as effective or even more so than using additional real data, obtaining a 0.63\% increase in performance with pix2pixHD data. On the other hand, the best-performing unsupervised model data proved as effective as additional real data, while the worst performance was achieved with CUT data, which is on par with not adding any data to the training set. With reference to the screening task, we observed that data synthesized with unsupervised methods proved slightly less effective than supplementary real data while still having a 0.44\% advantage over not using additional data. On the contrary, data generated with pix2pixHD and pix2pix was detrimental to model performance, with a performance degradation of at least 0.54 \% compared to using no additional data.

In all cases, we observed that CycleGAN data performs better than CUT data. As such, the cycle consistency mechanism for preserving content and transferring style is shown to be better suited for presentation attack generation than the patch-wise contrastive loss used by CUT. 

On the other hand, we also observe that pix2pixHD data is more effective than pix2pix data for PAD predictive performance despite producing less visually appealing presentations. This may be due to the distortions' positive regularising effect, given the low variability in the training data, although further research is needed to validate these claims. 

We also analyzed the quality of the generated images using the FID metric. The results reveal that CycleGAN produces the most faithful images, followed by CUT, pix2pixHD and finally pix2pix. This stands in contrast to the results observed in the print task experiments, where unsupervised models produced less effective data than supervised models. However, increased image quality is observed to be aligned with the screen task results. We attribute these differences in part to shortcomings in the image alignment process used for creating paired samples for supervised generative model training, which results in noisy generated samples. We hypothesize that this noise correlates well with the noise inherent to the paper texture while proving harmful for generating screen samples since the alignment process can interfere with the fine grain texture and moir{é} patterns expected in screen displays.

This article focused primarily on GAN-based unimodal, two-domain image-to-image translation models. Future work includes exploring the effectiveness of multi-modal models, as well as models that generate images in more than one domain. Additionally, we plan to complement our work by analyzing the quality of ID cards generated with recently proposed diffusion models. 

\section*{Acknowledgement}
The authors are grateful to Instituto Tecnológico de Informática. This work was supported by European Union (EU) Next-Generation, Plan de Recuperación, Transformación y Resiliencia through Convocatorias de ayudas 2021 Nº C005/21-ED, Proyecto Red.es and the German Federal Ministry of Education and Research and the Hessen State Ministry for Higher Education, Research and the Arts within their joint support of the National Research Center for Applied Cybersecurity ATHENE.

\bibliographystyle{IEEEtranN}
\bibliography{references.bib}

\vspace{+7cm}
\begin{IEEEbiography}[{\includegraphics[width=1in,height=1.25in,clip,keepaspectratio]{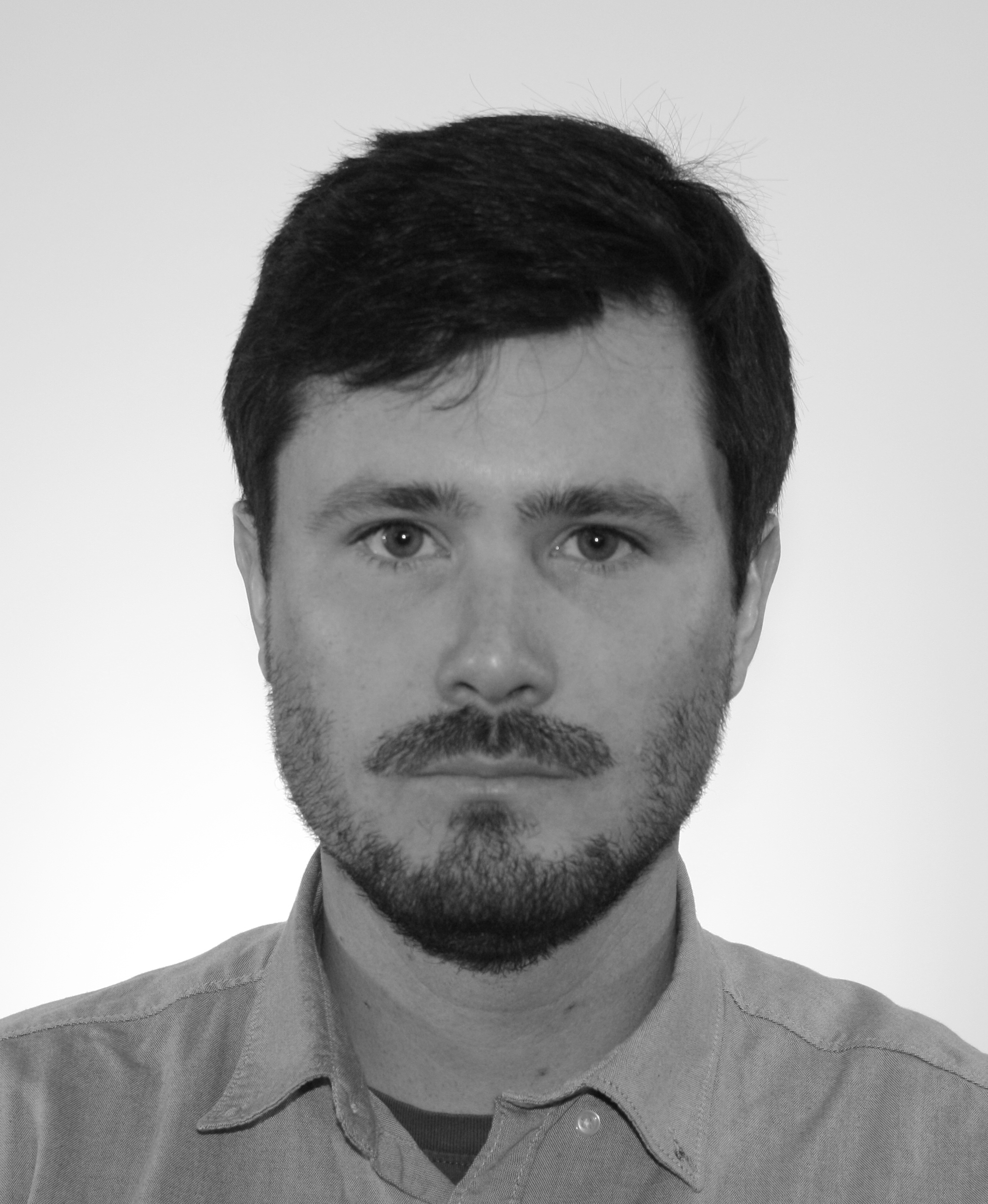}}]{Reuben Markham} received a B.Sc. degree in Mathematics from the University of Alicante in 2015, and a M.S. in Statistical and Computational Information Processing from the Complutense University of Madrid in 2016. At present, he works as a data scientist at the Instituto Tecnológico de Informática, where he has been involved in different client and research projects encompassing natural language processing, time series analysis and computer vision. His main interests include topics such as anomaly detection, probabilistic models and applied research.
\end{IEEEbiography}
\vspace{-0.3cm}
\begin{IEEEbiography}[{\includegraphics[width=1in,height=1.25in,clip,keepaspectratio]{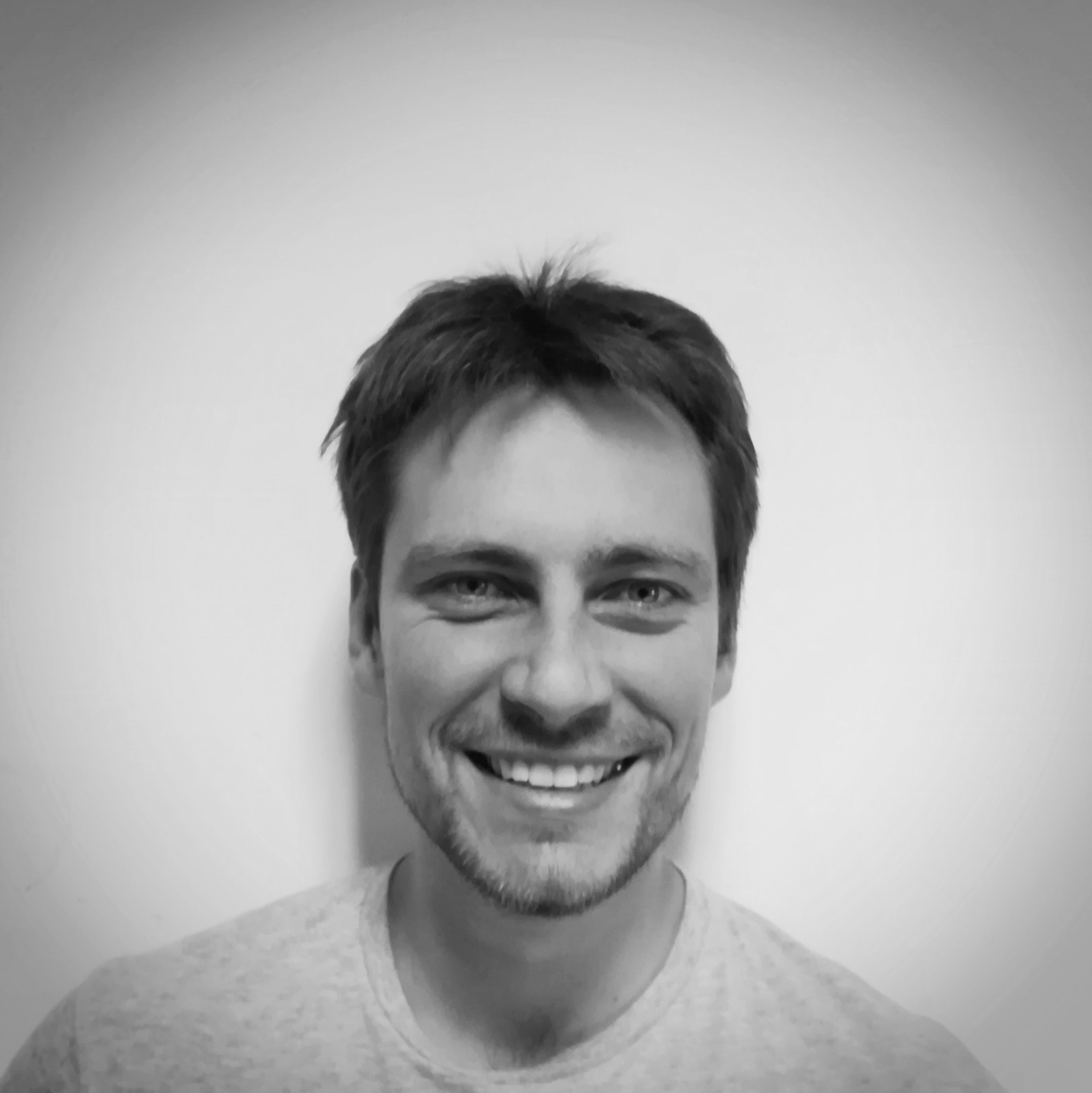}}]{Juan M. Espín López} received a B.Sc. degree in Mathematics from the University of Murcia in 2014, and a M.Sc. degree in Applied Mathematics in 2015. He is pursuing his PhD in Computer Science at the University of Murcia and works as a Senior Machine Learning Researcher at Facephi. His research interests focus on anti-spoofing systems for documents, face and voice, Continuous Authentication, speaker recognition, facial recognition, and machine learning and deep learning applications to the previous fields.
\end{IEEEbiography}
\vspace{-0.3cm}
\begin{IEEEbiography}
[{\includegraphics[width=1in,height=1.25in,clip,keepaspectratio]{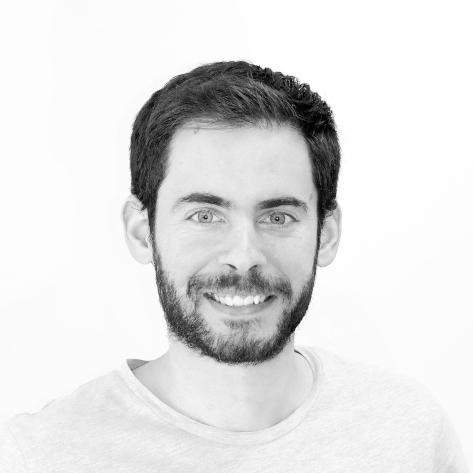}}]{Mario Nieto-Hidalgo} received a B.Sc degree in Computer Engineering in 2011, a M.Sc degree in Technologies of Information Society in 2012, and a PhD in Computer Science at the University of Alicante in 2017. Currently working for Facephi as a Senior Researcher and as Adjunct Professor at the University of Alicante. His main research interests are computer vision and machine learning for face recognition and ID documents, presentation attack systems, gait analysis and ambient assisted living.
\end{IEEEbiography}
\vspace{-0.3cm}

\begin{IEEEbiography}[{\includegraphics[width=1in,height=1.25in,clip,keepaspectratio]{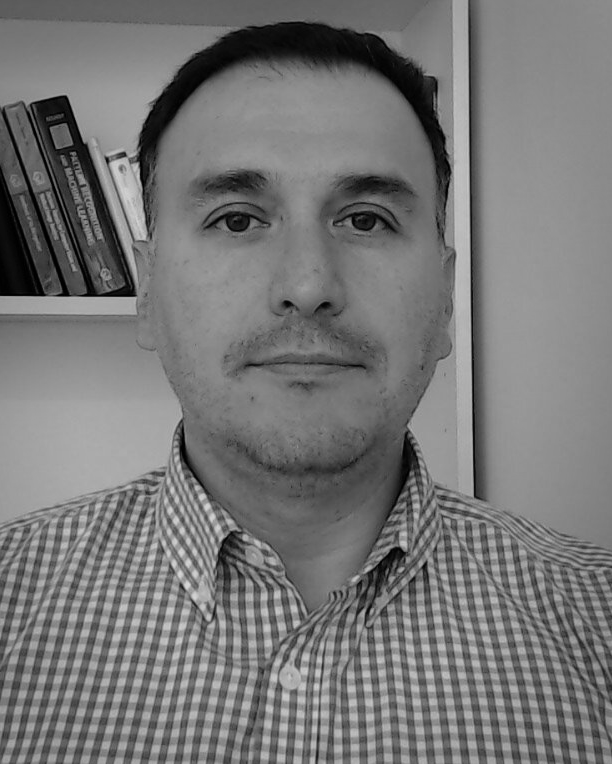}}]{Juan E. Tapia} received a P.E. degree in Electronics Engineering from Universidad Mayor in 2004, an M.S. in Electrical Engineering from Universidad de Chile in 2012, and a PhD from the Department of Electrical Engineering, Universidad de Chile in 2016. In addition, he spent one year of internship at the University of Notre Dame. In 2016, he received the award for best PhD thesis. From 2016 to 2017, he was an Assistant Professor at Universidad Andres Bello. From 2018 to 2020, he was the R\&D Director for the area of Electricity and Electronics at Universidad Tecnologica de Chile - INACAP and R\&D Director of TOC Biometrics up to November 2022. He is currently a biometric advisor and Senior Researcher at Hochschule Darmstadt (HDA). His main research interests include pattern recognition and deep learning applied to iris biometrics, morphing, feature fusion, and feature selection.
\end{IEEEbiography}

\end{document}